\documentclass[journal,10pt]{IEEEtran}
\usepackage{amssymb}
\usepackage{graphicx,psfrag} 
\usepackage{epsfig} 
\usepackage{esdiff} 
\usepackage{amsmath} 
\usepackage{xfrac}   
\usepackage{subfigure}  
\usepackage{pst-sigsys} 
\usepackage{algorithm}
\usepackage{cite}
\usepackage{algorithmic}
\usepackage{setspace}
\usepackage{color}

\usepackage{caption}
\usepackage{marginnote}
\usepackage{framed}
\usepackage{soul}
\DeclareCaptionLabelFormat{lc}{\MakeLowercase{#1}~#2}
\ifCLASSINFOpdf
\else
\fi
\begin{document}
\title{$\ell_1$-K-SVD: A Robust Dictionary Learning Algorithm With Simultaneous Update}
\author{Subhadip~Mukherjee, Rupam~Basu, and~Chandra~Sekhar~Seelamantula,~\IEEEmembership{Senior~Member,~IEEE}
\thanks{The authors are with the Department
of Electrical Engineering, Indian Institute of Science, Bangalore 560012, India. Emails: subhadip@ee.iisc.ernet.in,  rbasu28@ece.iisc.ernet.in, chandra.sekhar@ieee.org. 
}}
\markboth{}%
{Shell \MakeLowercase{\textit{et al.}}: Dictionary Learning Algorithm With Restricted Isometry Constraint}

\maketitle


\begin{abstract}
We develop a dictionary learning algorithm by minimizing the $\ell_1$ distortion metric on the data term, which is known to be robust for non-Gaussian noise contamination. The proposed algorithm exploits the idea of iterative minimization of weighted $\ell_2$ error. We refer to this algorithm as $\ell_1$-K-SVD, where the dictionary atoms and the corresponding sparse coefficients are simultaneously updated to minimize the $\ell_1$ objective, resulting in noise-robustness. We demonstrate through experiments that the $\ell_1$-K-SVD algorithm results in higher atom recovery rate compared with the K-SVD and the robust dictionary learning (RDL) algorithm proposed by Lu et al., both in Gaussian and non-Gaussian noise conditions. We also show that, for fixed values of sparsity, number of dictionary atoms, and data-dimension, the $\ell_1$-K-SVD algorithm outperforms the K-SVD and RDL algorithms when the training set available is small. We apply the proposed algorithm for denoising natural images corrupted by additive Gaussian and Laplacian noise. The images denoised using $\ell_1$-K-SVD are observed to have slightly higher peak signal-to-noise ratio (PSNR) over K-SVD for Laplacian noise, but the improvement in structural similarity index (SSIM) is significant (approximately $0.1$) for lower values of input PSNR, indicating the efficacy of the $\ell_1$ metric.

\end{abstract}

\IEEEpeerreviewmaketitle
\section{Introduction}
\IEEEPARstart{D}{ictionary} learning for sparse representation of signals has gained interest in the signal processing community in the past few years. A survey of the state-of-the-art techniques for dictionary learning is given in {\cite{Rubinstein,Tosic}} and references therein. Some seminal contributions to the problem of data-adaptive dictionary learning were made by Aharon et al. \cite{elad1,elad3}, who proposed the K-SVD algorithm, which, by far, is the most popular algorithm for dictionary design. Theoretical guarantees on the performance of K-SVD can be found in \cite{Arora,Schnass}. Dai et al. \cite{Dai} developed a generalized framework for dictionary learning, known as \textit{SimCO}, such that the MOD \cite{engan} and K-SVD emerge as special cases. K-SVD has been deployed to solve the problem of image denoising \cite{elad3}, specifically to suppress zero-mean additive white Gaussian noise. The principal idea behind the approach is to train a dictionary that is capable of representing image patches parsimoniously. Two training options have been proposed: (i) training on a corpus of clean image patches; and (ii) training on the noisy input image. Since the K-SVD algorithm is not suitable for handling large image patches due to computational overhead, addition of a global prior was proposed to enforce sparsity. The K-SVD uses the $\ell_2$ distortion as a measure of data fidelity. The dictionary learning problem has also been addressed from the analysis perspective \cite{rubinstein_peleg,eksioglu,dai_analysis}. Apart from denoising, dictionary-based techniques find applications in image super-resolution, inpainting, etc. \cite{yang,Sujit,Julien}.

\indent In many applications, the assumption of Gaussianness of noise may not be accurate, thus rendering the $\ell_2$-minimization based algorithms suboptimal. Moreover, algorithms based on $\ell_2$-error minimization may lead to over-smoothing of the image, causing loss of detail. One way to alleviate this problem is to develop algorithms to minimize the $\ell_1$ data error. We develop a dictionary learning algorithm that approximates the solution to the $\ell_1$ minimization problem by iteratively solving weighted $\ell_2$ minimization problems, known as {\it iteratively re-weighted least squares} (IRLS) \cite{daubechies,miosso,Endra}. We refer to the new algorithm as $\ell_1$-K-SVD, which is a robust dictionary learning algorithm that enables a simultaneous update of the dictionary and the corresponding coefficients over the current support.


\indent In dictionary learning for sparse coding, the standard metrics used for the data term and regularization are the $\ell_2$ and $\ell_1$ metrics, respectively. Algorithms for minimizing $\ell_1$-based data terms have been proposed in the context of non-negative matrix factorization (NMF) for document detection \cite{Kasiviswanathan}. Recently, the problem of robust dictionary learning has been addressed by Lu et al. \cite{Lu} in an online setting, where the training examples are revealed sequentially \cite{Mairal}. Unlike K-SVD, their algorithm does not offer the flexibility to simultaneously update the dictionary and the coefficients over the currently estimated support.
\subsubsection*{Our contribution}
We propose an algorithm for dictionary learning aimed at minimizing the $\ell_1$ distortion on the data term. The $\ell_1$ error is minimized using IRLS, using which we solve the $\ell_1$ minimization problem by iteratively solving a series of $\ell_2$ minimization problems (cf. Section \ref{prob_def_sec}). The use of $\ell_1$ metric for data fidelity results in robustness for suppression of impulsive noise from images while preserving structure, as demonstrated by the experimental results (cf. Section \ref{exp_results_sec}). Unlike \cite{Lu}, in $\ell_1$-K-SVD, the dictionary atoms and the entries of the coefficient matrix are simultaneously updated (similar to K-SVD), offering better convergence performance. 
\section{Proposed Dictionary Learning Algorithm}
\label{prob_def_sec}
Denote the training data by $\mathcal{T}$, which contains $N$ exemplars $\left\{ \bold y_n  \right\}_{n=1}^{N}$ in $\mathbb{R}^m$, corrupted by additive noise. The objective is to learn a dictionary $D$ containing $K$ atoms, tailored to $\mathcal{T}$, such that $D$ represents the members of $\mathcal{T}$ using sparse coefficient vectors $ \bold x_n  $. Typically, $D$ is over-complete, meaning that $m<K$. The symbols $Y$ and $X$ denote the matrices constructed by stacking the training examples $\bold y_n$ and the corresponding coefficients $\bold x_n$ in the columns. In order to achieve robustness to noise, we propose to solve the following optimization problem:
\begin{equation}
\underset{D, \bold x_n}{\min}\sum_{n=1}^{N} \left\| \bold y_n-D \bold x_n \right\|_1 +  \sum_{n=1}^{N}\lambda_n \left\| \bold x_n \right\|_1.
\label{prob_L1}
\end{equation}
The cost function in (\ref{prob_L1}) can be interpreted from a Bayesian perspective, where a Laplacian model is assumed for the prior as well as for the additive noise. The parameters $\lambda_n$ in the regularization term adjust the sparsity of the resulting representation versus data fidelity. To solve (\ref{prob_L1}), we adopt the alternating minimization strategy, wherein one starts with an initial guess for $D$, and updates $\mathbf{x}_n$ and $D$, alternately. 


\subsection{Sparse Coding}
\label{sparse_coding_sec}
To update the coefficient vectors $\mathbf{x}_n$ when $D$ is fixed, one needs to solve $N$ independent problems of the form 
\begin{equation}
\underset{\bold x_n}{\min}\left\| \bold y_n-\hat{D} \bold x_n \right\|_1 +  \lambda_n \left\| \bold x_n \right\|_1,
\label{sparse_coding}
\end{equation}
where $\hat{D}$ denotes the estimate of the dictionary in the current iteration. To solve (\ref{sparse_coding}) using IRLS, one has to start with an initial guess $\bold x_n^{(0)}$ and update $\bold x_n$ in the $k^{\text{th}}$ iteration as
\begin{equation*}
\bold x_n^{(k+1)}= \left( \hat{D}^T W_{1n}^{(k)}\hat{D}+\lambda_n W_{2n}^{(k)} \right)^{-1}\hat{D}^T W_{1n}^{(k)} \bold y_n,
\label{xn_update}
\end{equation*}
where $W_{1n}$ and $W_{2n}$ are diagonal weight matrices of appropriate dimensions. Typically, one begins the iterations by setting $W_{1n}^{(0)}$ and $W_{2n}^{(0)}$ to identity matrices, for all $n$, and updating in the $k^{\text{th}}$ iteration as
\small
\begin{eqnarray*}
W_{1n}^{(k+1)}(j) = \frac{1}{\left|\left( \bold y_n-\hat{D} \bold x_n^{(k)} \right)_j\right| + \epsilon},
W_{2n}^{(k+1)}(j) = \frac{1}{ \left|\left(\bold x_n^{(k)} \right)_j \right|+ \epsilon}, 
\end{eqnarray*}
\normalsize
where $W(j)$ denotes the $j^{\text{th}}$ diagonal entry of a diagonal matrix $W$, and $(\bold x)_j$ denotes the $j^{\text{th}}$ entry of the vector $\bold x$. A small positive constant $\epsilon$ is added to the denominator to ensure numerical stability. One can also choose to work with an equivalent constrained version of (\ref{sparse_coding}) of the form
\begin{equation}
\underset{\bold x_n}{\min}  \left\| \bold y_n-\hat{D} \bold x_n \right\|_1 \text{\,\,subject to\,\,}\left\| \bold x_n \right\|_1 \leq \tau_n,
\label{prob_L1_constrained}
\end{equation}
where $\tau_n$ controls the sparsity of the resulting solution, with smaller values of $\tau_n$ resulting in higher sparsity. Efficient optimization packages, such as CVX \cite{grant} and ManOpt \cite{manopt} are available to solve problems of the form given in (\ref{sparse_coding}) and (\ref{prob_L1_constrained}). To further reduce the sensitivity of the solution to noise, we apply an appropriate threshold $T_0$ on the entries of $X$, following sparse coding. The optimum value of the threshold is application specific, and the values chosen in our experiments are indicated in Section \ref{exp_results_sec}. Thus, the proposed approach may also be interpreted as $\ell_1$ minimization followed by pruning.
\vspace{-0.5in}
\begin{center}

\begin{figure}[t]
\begin{tabular}{cccc}
 \includegraphics[width=1.5in]{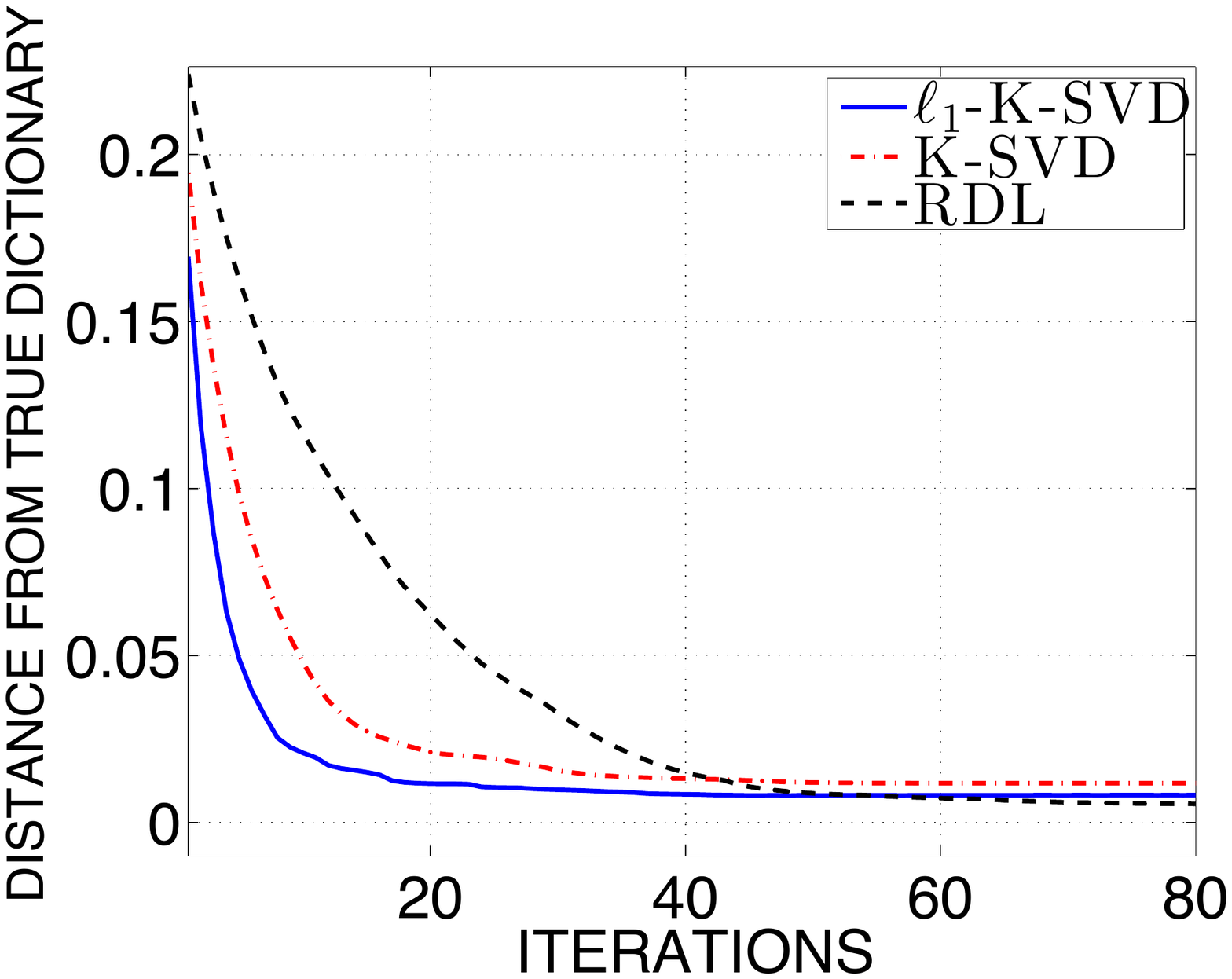} &
\includegraphics[width=1.5in]{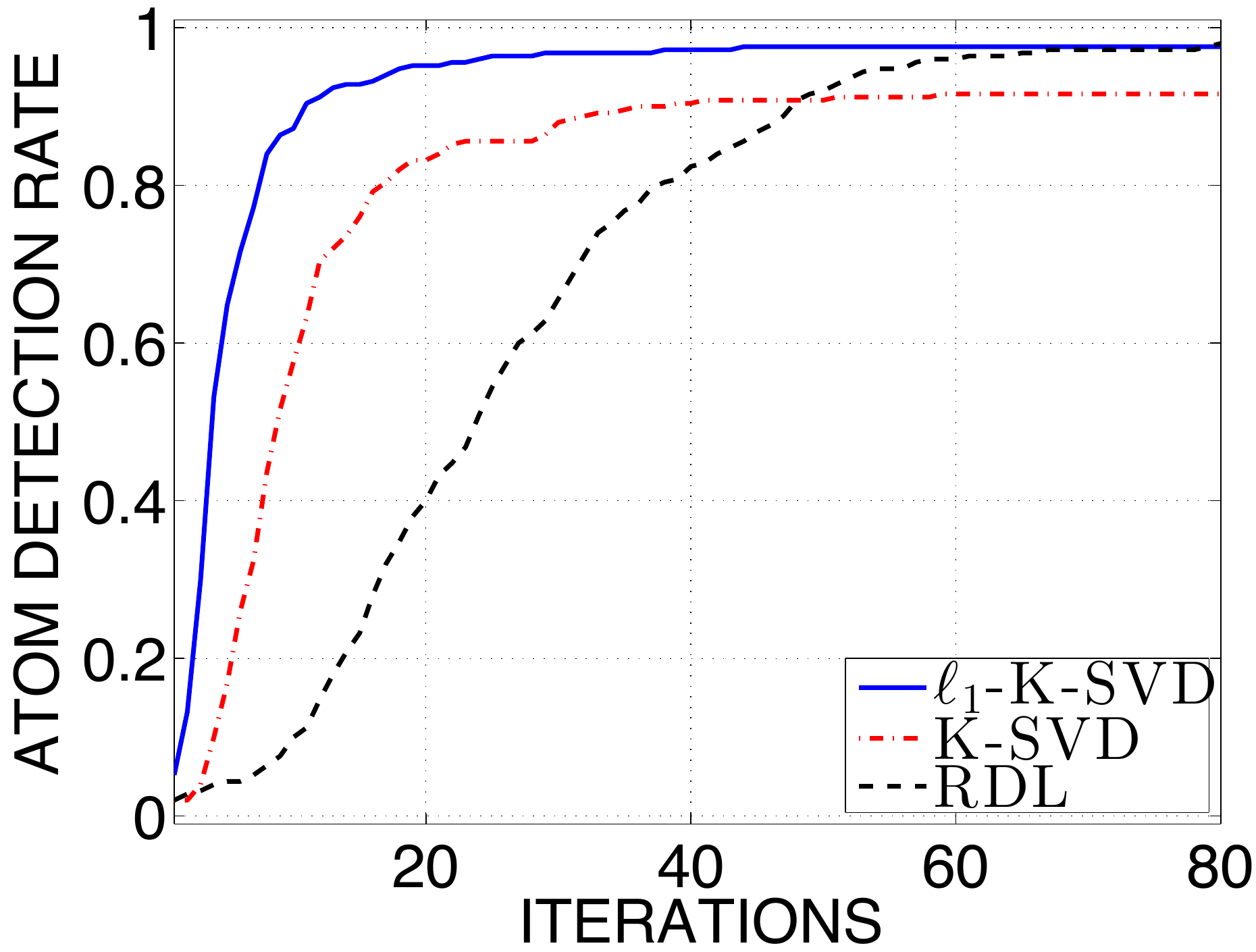}\\
\includegraphics[width=1.5in]{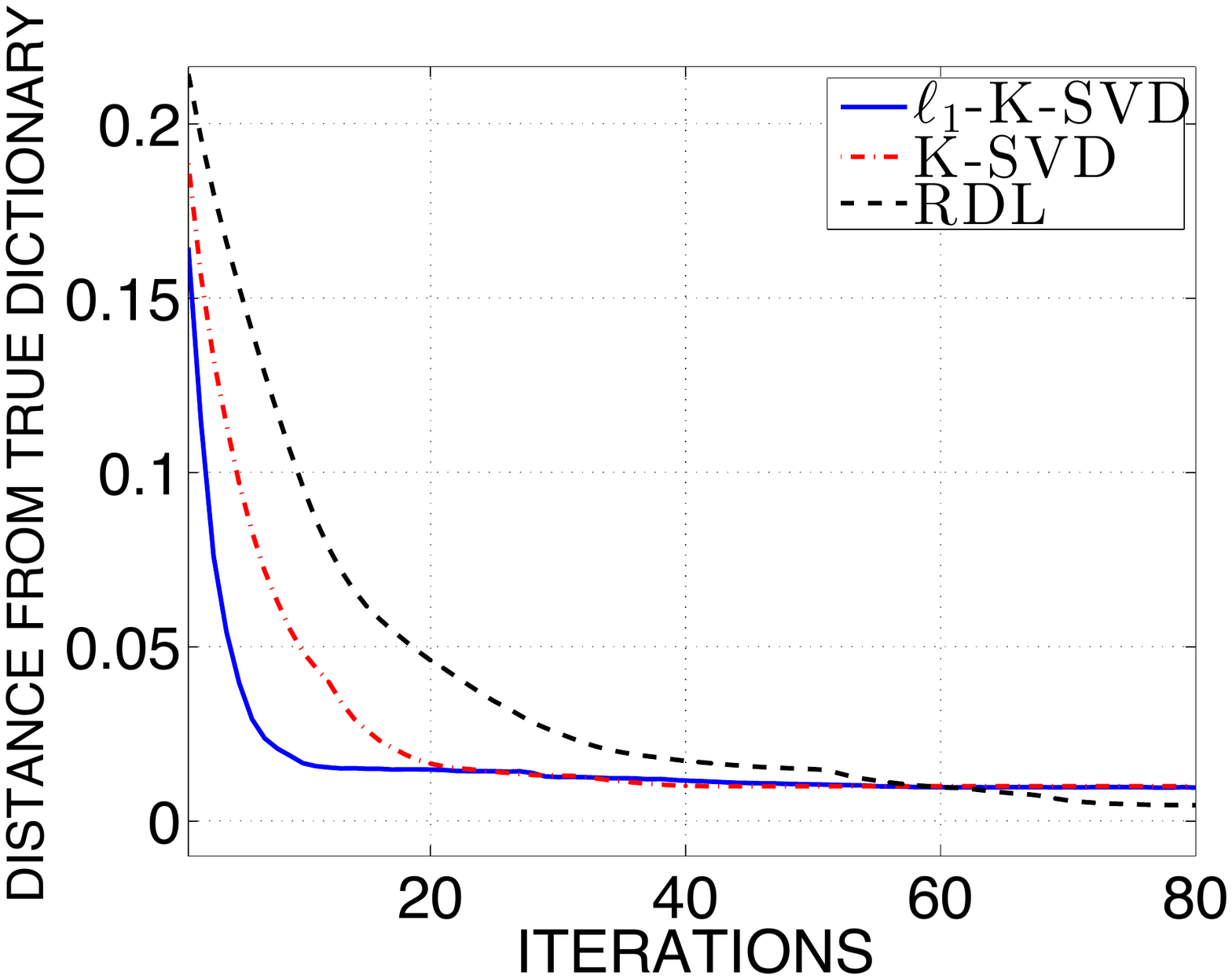}&
\includegraphics[width=1.5in]{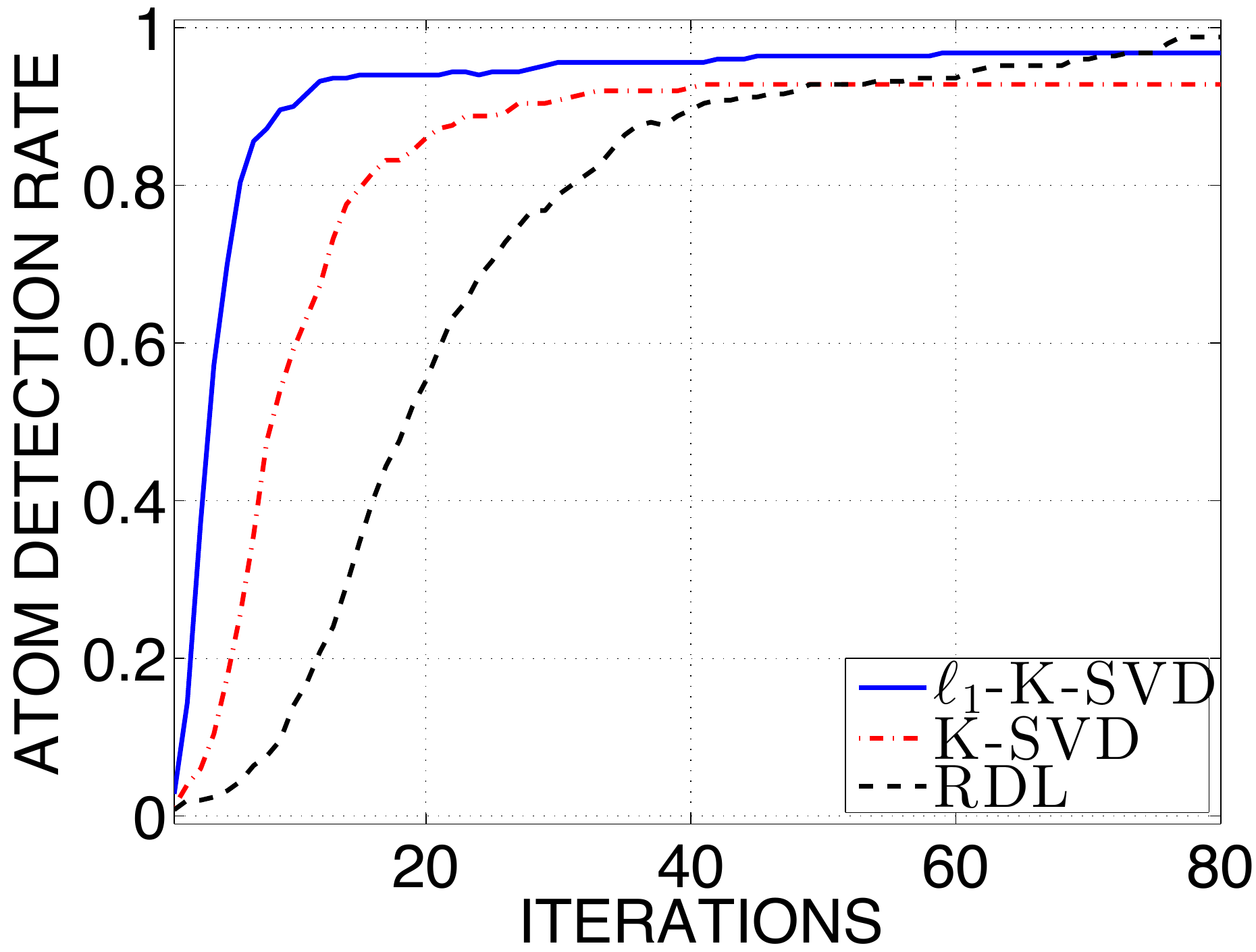}\\
\end{tabular}
\caption{\small{(Color online) Performance comparison of the $\ell_1$-K-SVD, K-SVD, and RDL algorithms in terms of ADR and distance metric $\kappa$, for Gaussian (first row) and Laplacian (second row) noise. The plots are averaged over five independent realizations. The input SNR is $20$ dB. The number of examples in the training set is $N=1500$}. \vspace{-0.4in}}
\label{comparison_fig}
\end{figure}
\end{center}

\begin{center}
\begin{figure}[h]
	\begin{tabular}{ccccc}
		\includegraphics[width=1.in]{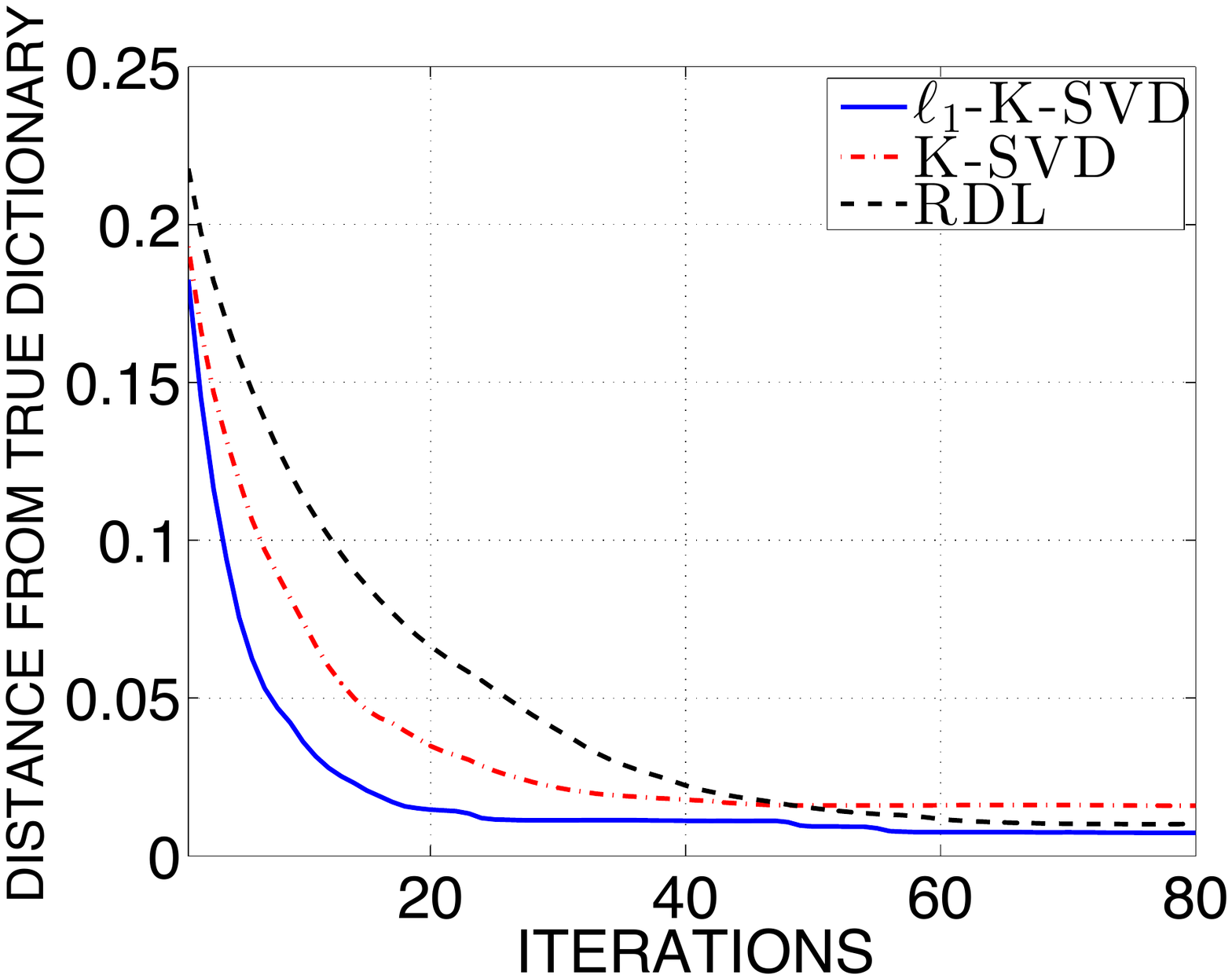}&
		\includegraphics[width=1.in]{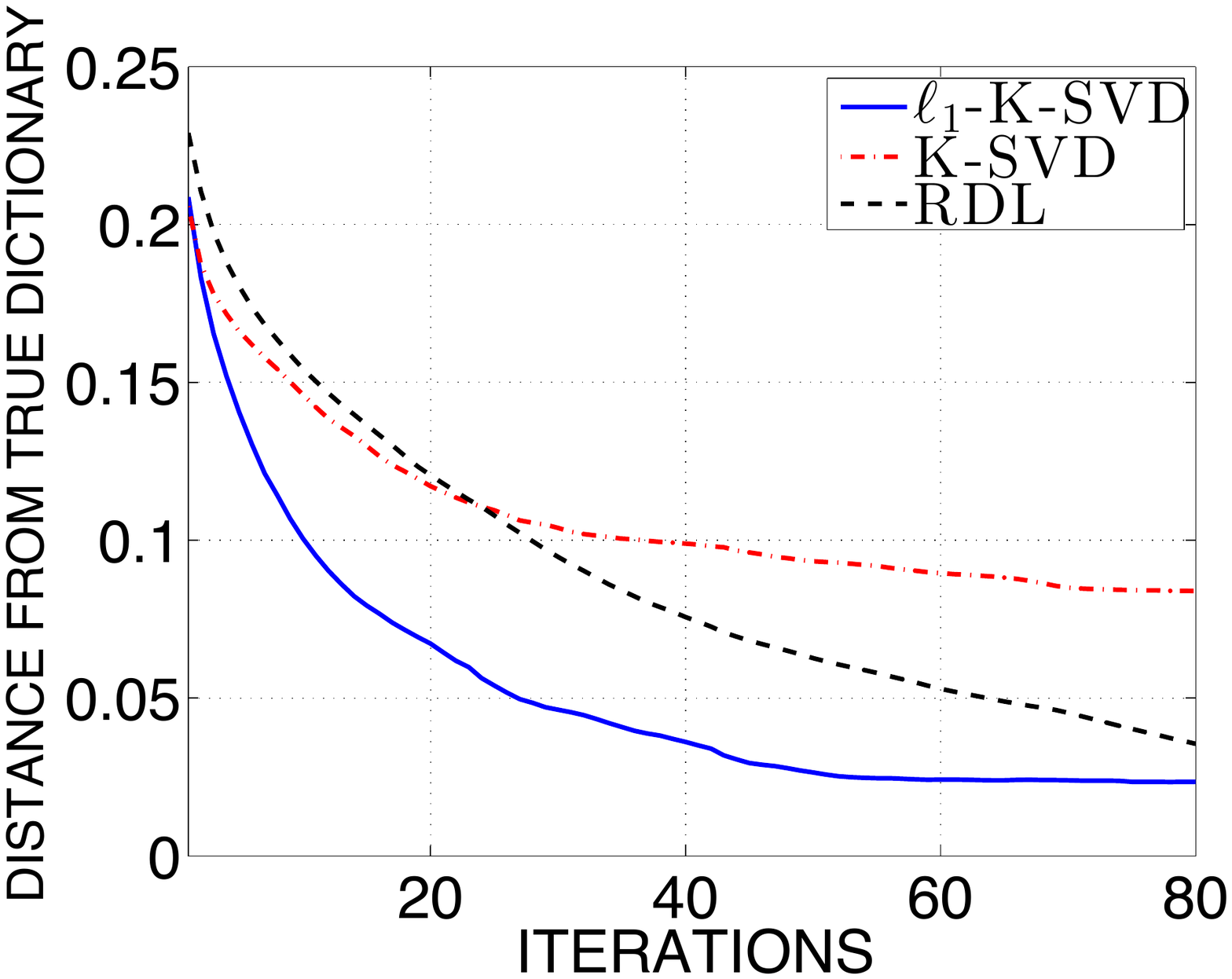}&
		\includegraphics[width=1.in]{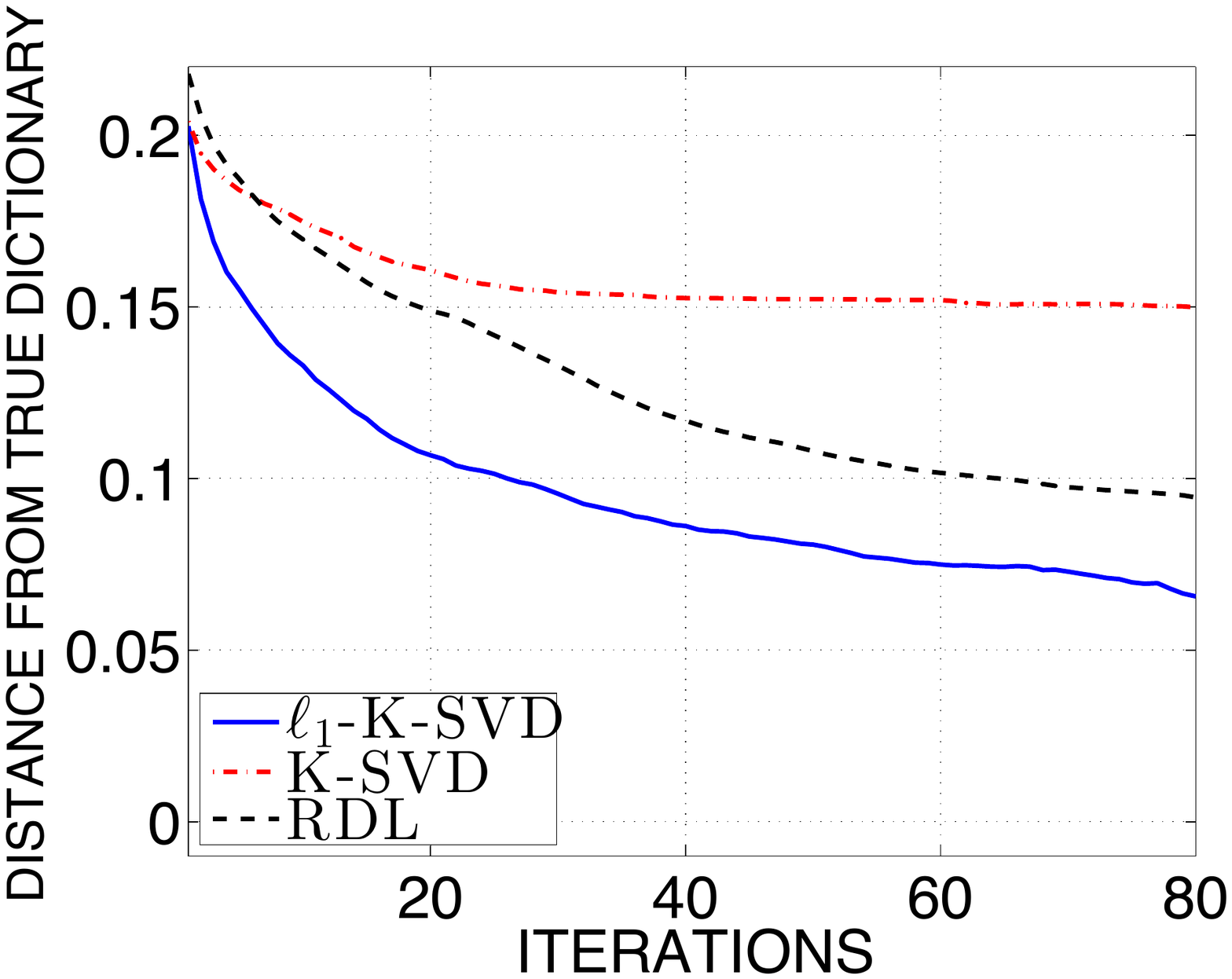}\\
		\includegraphics[width=1.in]{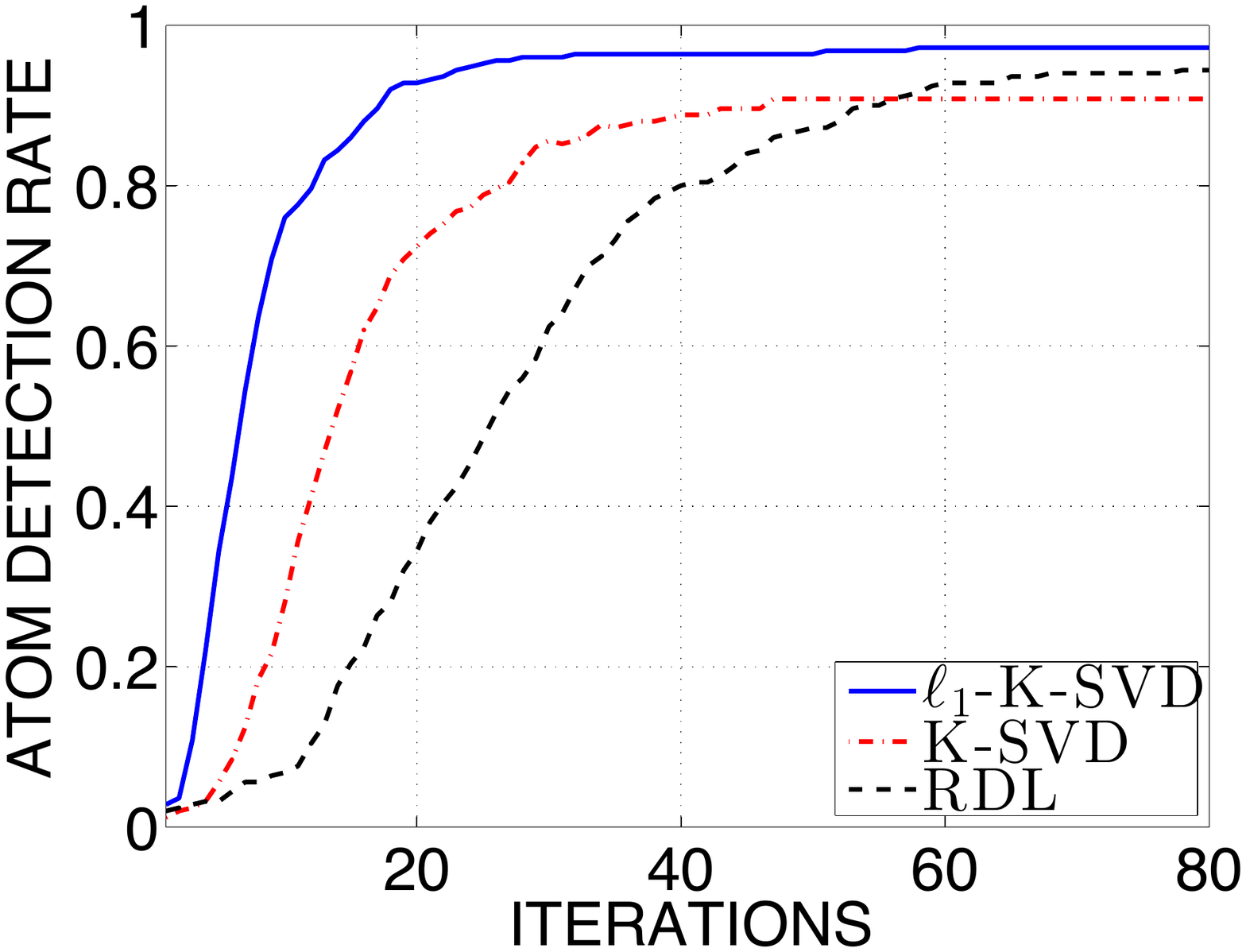}&
		\includegraphics[width=1.in]{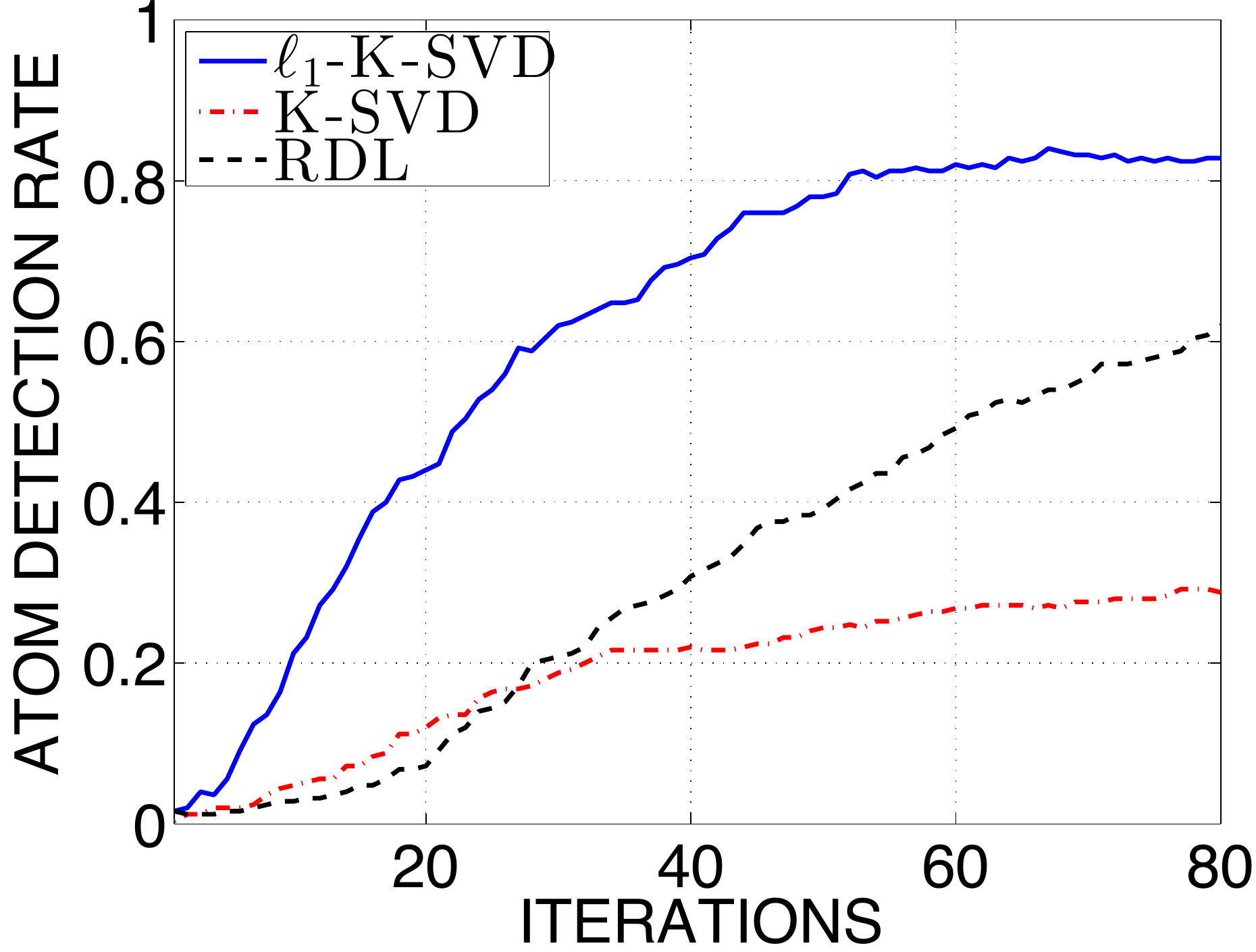}&
		\includegraphics[width=1.in]{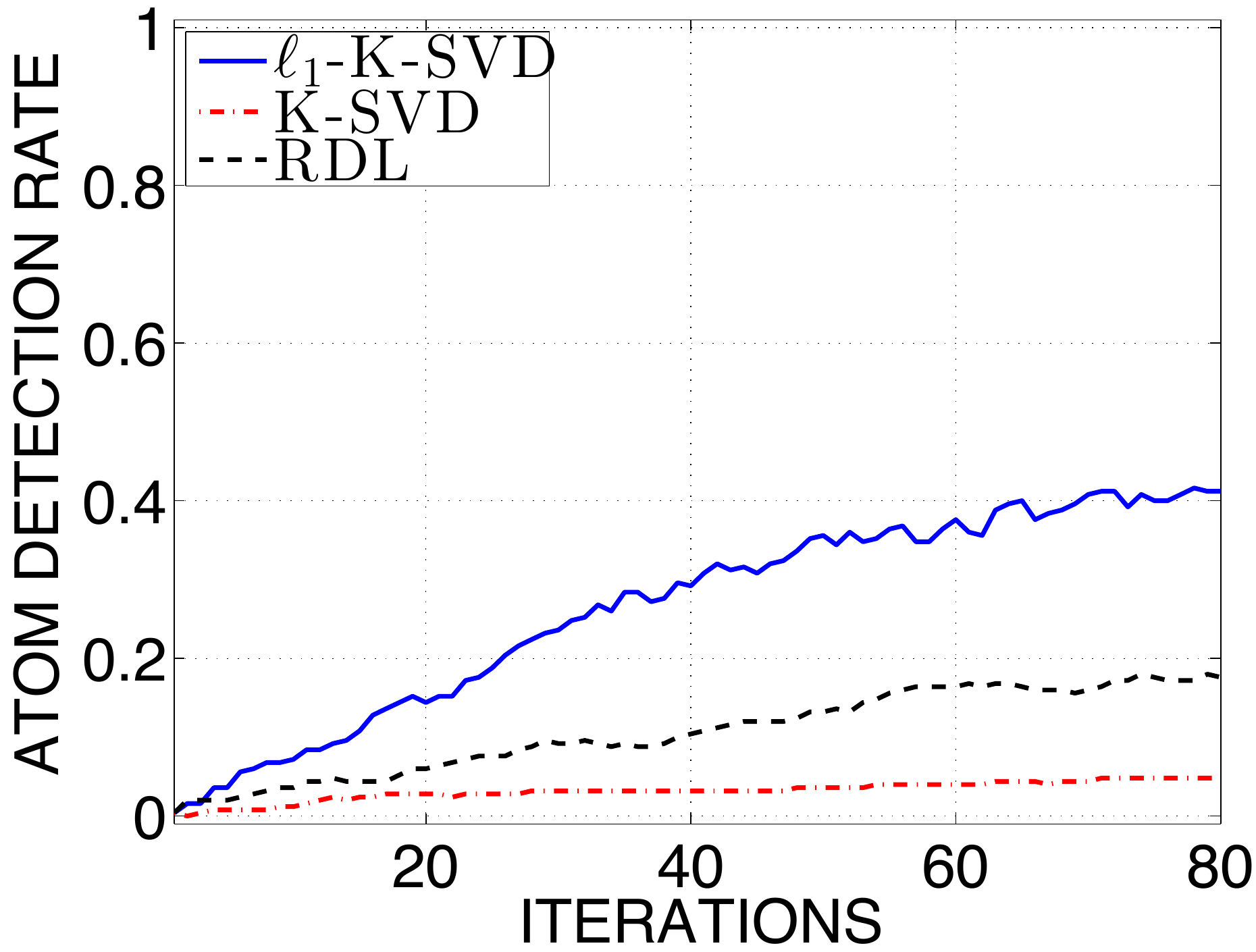}\\
		$N=600$ &  $N=300$ &  $N=200$\\
	\end{tabular}
	\caption{\small{(Color online) Performance comparison of $\ell_1$-K-SVD, K-SVD, and RDL algorithms for varying data-size $N$. The values of $m$, $s$, and $K$ are kept fixed at $m=20$, $s=3$, and $K=50$. The figures in the first and second rows show the variations of the distance metric $\kappa$ and the ADR, respectively, with iterations. The dataset is corrupted by additive Laplacian noise with SNR $20$ dB.}\vspace{-.4in}}
	\label{EffectOfdataSize_fig}
\end{figure}
\end{center}
\begin{algorithm}[t]
	\caption{{\bf ($\ell_1$-K-SVD):} To find $\bold u$ and $\bold v$ such that $\left\| E-\bold u \bold v^T \right\|_1$ is minimized, subject to $\left\| \bold u  \right\|_2=1$.}
	\begin{algorithmic}
		\STATE {\bf 1.} {\bf Input:} Error matrix $E$, number of iterations $J$.
		\vspace{0.02in}
		\STATE {\bf  2.} {\bf Initialization:} 
		\begin{itemize}
			\item Set the iteration counter $p\leftarrow 0$, $\bold u^{(p)} \leftarrow \bold a$, and $\bold v^{(p)} \leftarrow \sigma\bold b$, where $\bold a$ and $\bold b$ are the left and right singular vectors, respectively, corresponding to the largest singular value $\sigma$ of $E$. 
			\item Initialize the weight matrices $W_n^{(p)} \leftarrow I$, for $n=1,2,\cdots,M$, where $I$ denotes the identity matrix and $M$ is the number of columns in $E$.
		\end{itemize}
		
		\vspace{0.02in}
		\STATE {\bf  3.} {\bf Iterate the following steps $J$ times:} 
		\begin{itemize}
			\item $w_n^{(p+1)}(j)\leftarrow \frac{1}{\left|\left( \bold e_n- v_n^{(p)}\bold u^{(p)} \right)_j\right|+\epsilon}$
			\small
			\item $\bold u^{(p+1)} \leftarrow \displaystyle  \left[\sum_{n=1}^{M}{\left(v_n^{(p)}\right)}^2 W_n^{(p+1)} \right]^{-1} \left[\sum_{n=1}^{M} v_n^{(p)} W_n^{(p+1)} \bold e_n \right]$
			\normalsize
			\item $v_n^{(p+1)} \leftarrow \frac{\left(\bold u^{(p+1)}\right)^T W_n \bold e_n}{\left(\bold u^{(p+1)}\right)^T W_n \left(\bold u^{(p+1)}\right)}$, for $n=1,2,\cdots,M$
			\item $p \leftarrow p+1$
		\end{itemize}
		\STATE {\bf  4.} {\bf Scaling:} Set $\bold u \leftarrow \frac{\bold u^{(p)}}{\left\| \bold u^{(p)} \right\|_2}$ and $\bold v \leftarrow \left\| \bold u^{(p)} \right\|_2 \bold v^{(p)}$
		\STATE {\bf  5.} {\bf Output:} Solution $\left(\bold u, \bold v \right)$ to (\ref{prob_L}).
	\end{algorithmic}
\end{algorithm}



\begin{table}[t]
\centering 

\begin{tabular}{|c| c|  c| }

\hline 
$\sigma=15$  & $\sigma=25$ & $\sigma=35$  \\ 
\hline 

 $\lambda=1,\textcolor{red}{1}$  & $\lambda=1,\textcolor{red}{1}$ &  $\lambda=8,\textcolor{red}{8}$  \\ 
  $n_{p}=0.18,\textcolor{red}{0.15}$  & $n_{p}=0.08,\textcolor{red}{0.05}$  &  $n_{p}=0.05,\textcolor{red}{0.04}$  \\  
\hline 
\end{tabular}
\vspace{0.1cm}
\caption{\small{(Color online) Values of the regularization parameter $\lambda$ and the fraction of entries $n_p$ retained in the thresholding step of $\ell_1$-K-SVD, in case of Laplacian (black) and Gaussian (red) noise.}\vspace{-.2in}}
\label{table_params} 
\end{table}

\begin{table*}[t]
\centering 

\begin{tabular}{|c| c c c| c c  c|}
\hline
&  & K-SVD  &  & &$\ell_1$-K-SVD&\\
\hline 
 &$\sigma=15$  & $\sigma=25$ & $\sigma=35$ & $\sigma=15$ & $\sigma=25$  & $\sigma=35$ \\ 
\hline 
Gaussian noise & $31.72/\textcolor{red}{31.23}$  & $28.66/\textcolor{red}{28.26}$  &  $26.43/\textcolor{red}{26.18}$   & $30.77/\textcolor{red}{30.08}$   &   $28.58/\textcolor{red}{28.00}$     & $25.98/\textcolor{red}{26.22}$  \\ 
& $0.80/\textcolor{red}{0.81}$    &  $0.68/\textcolor{red}{0.67}$  &  $0.56/\textcolor{red}{0.54}$   &  $0.78/\textcolor{red}{0.77}$  &   $0.73/\textcolor{red}{0.73}$     & $0.62/\textcolor{red}{0.61}$    \\ 
 \hline

Laplacian noise & $31.50/\textcolor{red}{31.04}$  & $28.39/\textcolor{red}{27.98}$ & $26.12/\textcolor{red}{25.86}$ & $31.70/\textcolor{red}{30.96}$   & $28.46/\textcolor{red}{28.30}$ & $26.79/\textcolor{red}{26.55}$\\ 
 &$0.79/\textcolor{red}{0.80}$ & $0.67/\textcolor{red}{0.66}$  & $0.55/\textcolor{red}{0.53}$ & $0.81/\textcolor{red}{0.80}$   &  $0.76/\textcolor{red}{0.75}$ & $0.66/\textcolor{red}{0.60}$\\ 
\hline 
\end{tabular}
\vspace{0.1cm}
\caption{\small{(Color online) Comparison of output PSNR (DB) and SSIM (corresponding to the first and second rows in each cell, respectively) for the K-SVD and $\ell_1$-K-SVD algorithms on the \textit{House} (black) and \textit{Tower} (red) images corrupted by additive Gaussian and Laplacian noise, for different input noise levels. The values reported are obtained by averaging over five independent realizations. The values of PSNR and SSIM did not change much from one realization to another}.\vspace{-.2in}}
\label{table_image_gauss_laplacian} 
\end{table*}

\vspace{-.2in}
\subsection{Dictionary Update}
\indent Similar to the K-SVD algorithm, we adopt a sequential updating strategy for the dictionary atoms $\left\{\bold d_i\right\}_{i=1}^{K}$. To simultaneously update the $j^{\text{th}}$ atom in the dictionary and the corresponding row in $X$, we focus on the error matrix $E_j = Y-\sum_{i\neq j}\bold d_i \bold x^i$, resulting from the absence of $\bold d_j$, where $\bold x^i$ denotes the $i^{\text{th}}$ row of $X$. In order to obtain a refined estimate of $\bold d_j$ and $\bold x^j$, one is required to solve
\small
\begin{eqnarray}
\underset{\bold d_j, \bold x^j}{\min} \left\| E_j-\bold d_j \bold x^j \right\|_1 \text{\,\,s.t.\,\,}\mathcal{S}\left( \bold x^j \right)=\Omega_j \text{\,\,and\,\,} \left\| \bold d_j\right\|_2=1,
\label{RDL_eqn}
\end{eqnarray}
\normalsize
where $\left\| A \right\|_1$ denotes the sum of the absolute values of the entries of matrix $A$. We seek to minimize the cost function subject to the constraint that $\mathcal{S}\left( \bold x^j \right)=\Omega_j$, where $\mathcal{S}(\cdot)$ denotes the support operator, defined as $\mathcal{S}\left( \bold z \right)=\left\{ i:\mathbf{z}_i \neq 0 \right\}$. Therefore, it suffices to consider only those columns of $E_j$ whose indices are in the support set $\Omega_j$, denoted by $E_j \big|_{\Omega_j}$. Furthermore, to get rid of scaling ambiguity, we restrict $D$ to have unit-length atoms, that is, $\left\| \bold d_i \right\|_2=1$ for all $i$. Consequently, solving (\ref{RDL_eqn}) is equivalent to solving \small
\begin{eqnarray}
\underset{\bold u,\bold v}{\min}\left\| E-\bold u \bold v^T \right\|_1 =\underset{\bold u, \bold v}{\min}  \sum_{n=1}^{M} \left\| \bold e_n- v_n\bold u \right\|_1 \text{\,s.t.\,} \left\|\bold u\right\|_2 =1,
\label{prob_L}
\end{eqnarray}
\normalsize
where $\bold e_n$ is the $n^{\text{th}}$ column of $E$ and $v_n$ denotes the $n^{\text{th}}$ entry of the vector $\bold v$. The key idea behind the $\ell_1$-K-SVD approach is to approximate the $\ell_1$-norm using a reweighted $\ell_2$ metric: 
\begin{eqnarray*}
\sum_{n=1}^{M} \left\| \bold e_n- v_n\bold u \right\|_1 &\approx& \sum_{n=1}^{M} \left( \bold e_n- v_n\bold u \right)^T W_n \left( \bold e_n- v_n\bold u \right).
\end{eqnarray*}
Differentiating with respect to $\bold u$ and setting to zero, we get 
\begin{equation*}
\bold u =  \left(\sum_{n=1}^{M}v_n^2 W_n \right)^{-1} \left(\sum_{n=1}^{M} v_n W_n \bold e_n \right).
\end{equation*}
Similarly, for a fixed $\bold u$, we have $v_n=\frac{\bold u^T W_n \bold e_n}{\bold u^T W_n \bold u}$. Once the update of $\bold u$ and $\bold v$ is done, the $j^{\text{th}}$ diagonal element of $W$ should be updated as $w_n(j)\leftarrow \frac{1}{\left|\left( \bold e_n- v_n\bold u \right)_j\right|+\epsilon}$. Dictionary update is performed by setting $\bold d_j=\bold u_0$ and $\bold x^j \big|_{\Omega_j}=\bold v_0^T$, where $\left(\bold u_0,\bold v_0 \right)$ solves (\ref{prob_L}). The step-by-step description to obtain the solution of (\ref{prob_L}) is given in Algorithm 1. Experimentally we found that $J=10$ suffices for the convergence. The $\ell_1$-K-SVD algorithm is computationally more expensive than the K-SVD and starts with the K-SVD initialization. However, the quality of the solution obtained using $\ell_1$-K-SVD is better than that obtained using K-SVD, in the case where the training set is contaminated by non-Gaussian as well as Gaussian noise.


\section{Experimental Results}
\label{exp_results_sec}


\subsubsection{Synthesized Signal}
To validate the $\ell_1$-K-SVD algorithm, we first conduct experiments on synthesized training data, for Gaussian as well as Laplacian noise. Since the ground-truth dictionary $D$ is known in the experiments, we use two performance metrics, namely the atom detection rate (ADR) and distance $\kappa \left( \hat{D},D \right)$ of the estimated dictionary $\hat{D}$ from the ground-truth $D$. To compute ADR, we count the number of recovered atoms in $D$. An atom $\bold d_i$ in $D$ is considered to be recovered if $\left| \bold d_i^T \hat{\bold  d}_j \right|$ exceeds $0.99$ for some $j$. Since $D$ and $\hat{D}$ contain unit-length atoms, this criterion ensures a near-accurate atom recovery. Finally, to compute ADR, we take the ratio of the number of recovered atoms to the total number of atoms. The metric $\kappa$ measures the closeness of $\hat{D}$ to $D$, and is defined as $\kappa =  \frac{1}{K}\sum_{i=1}^{K}\underset{1 \leq j \leq K}{\min} \left(1-\left| \bold d_i^T \hat{\bold d}_j \right| \right)$. The value of $\kappa$ satisfies $0<\kappa<1$. If $\kappa$ is close to $0$, it indicates that the recovered dictionary closely matches with the ground-truth. The experimental setup is identical to that in \cite{elad1}. The ground-truth dictionary is created by randomly generating $K=50$ vectors in $\mathbb{R}^{20}$, followed by column normalization. Each column of the coefficient matrix $X$ has $s=3$ non-zeros, with their locations chosen uniformly at random and amplitudes drawn from the $\mathcal{N}(0,1)$ distribution. Subsequently, $D$ is combined with $X$ and contaminated by additive noise, to generate $N=1500$ exemplars. All algorithms are initialized with the training examples and the iterations are repeated $80$ times, as suggested in \cite{elad1}. The optimum choice of the hard-threshold $T_0$, found experimentally, is given by $T_0=0.03\|X\|_F$. The variation of $\kappa$ and ADR with iterations, averaged over five independent trials, is demonstrated in Fig.~\ref{comparison_fig}. To facilitate fair comparison of $\ell_1$-K-SVD with the K-SVD and RDL \cite{Lu}, true values of the $\ell_1$ and $\ell_0$ norm of the ground-truth coefficient vectors, that is, true values of $\tau_n$ in (\ref{prob_L1_constrained}) and the parameter $s$, respectively, are supplied to the algorithms. OMP is used for sparse coding in K-SVD. The $\ell_1$-K-SVD algorithm, as one can see from Fig.~\ref{comparison_fig}, results in a superior performance over RDL and its $\ell_2$-based counterpart K-SVD, both in terms of $\kappa$ and ADR. One can observe similar robustness of the $\ell_1$-K-SVD algorithm to both Gaussian and Laplacian noise from the experimental results. The faster convergence of the $\ell_1$-K-SVD algorithm compared to RDL is attributed to the simultaneous update of the dictionary and the coefficients. However, they result in similar performance after sufficiently many iterations. The execution times taken for each iteration of the K-SVD, RDL and $\ell_1$-K-SVD are $0.53$, $45.53$, and $32.90$ seconds, respectively, when executed on a MATLAB 2011 platform, running on a Macintosh OSX system with $8$ GB RAM and $3.2$ GHz core-i5 processor.   

\indent To analyze the effect of the size of the training set on the performance, we decrease the value of $N$, keeping $m$, $s$, and $K$ fixed. From the plots in Fig. \ref{EffectOfdataSize_fig}, we observe that the $\ell_1$-K-SVD algorithm is superior to RDL and K-SVD, when the value of $N$ is small. For $N=200$, we note that the $\ell_1$-K-SVD algorithm recovers $40\%$ of the atoms accurately, whereas the ADR for RDL and K-SVD is below $20\%$. This experiment suggests that the $\ell_1$-K-SVD is particularly suitable when the number of training examples is small.

\subsubsection{Application to Image Denoising}
\begin{center}
\begin{figure}[t]
	\begin{tabular}{ccccc}
		\includegraphics[width=1.5in]{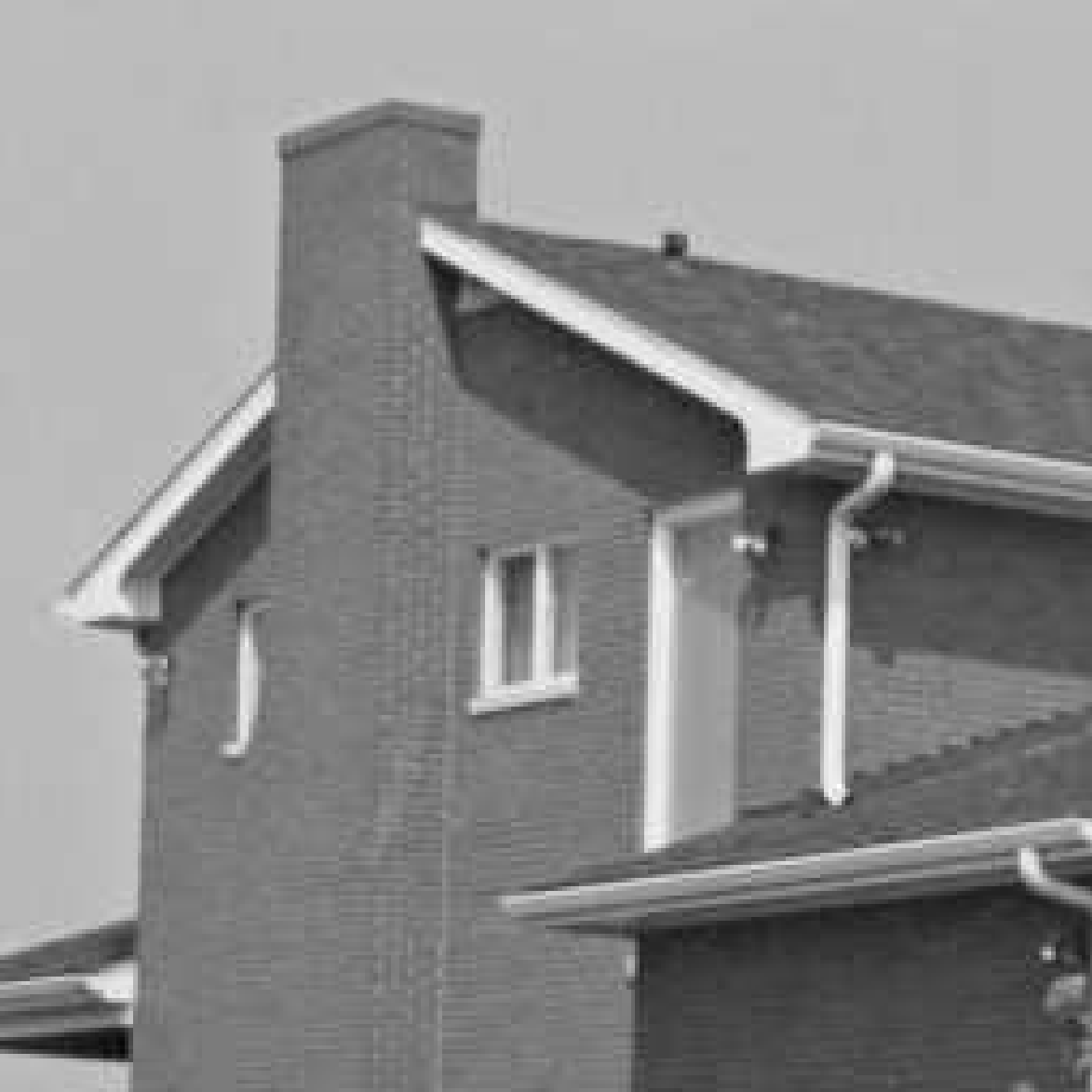}&
		\includegraphics[width=1.5in]{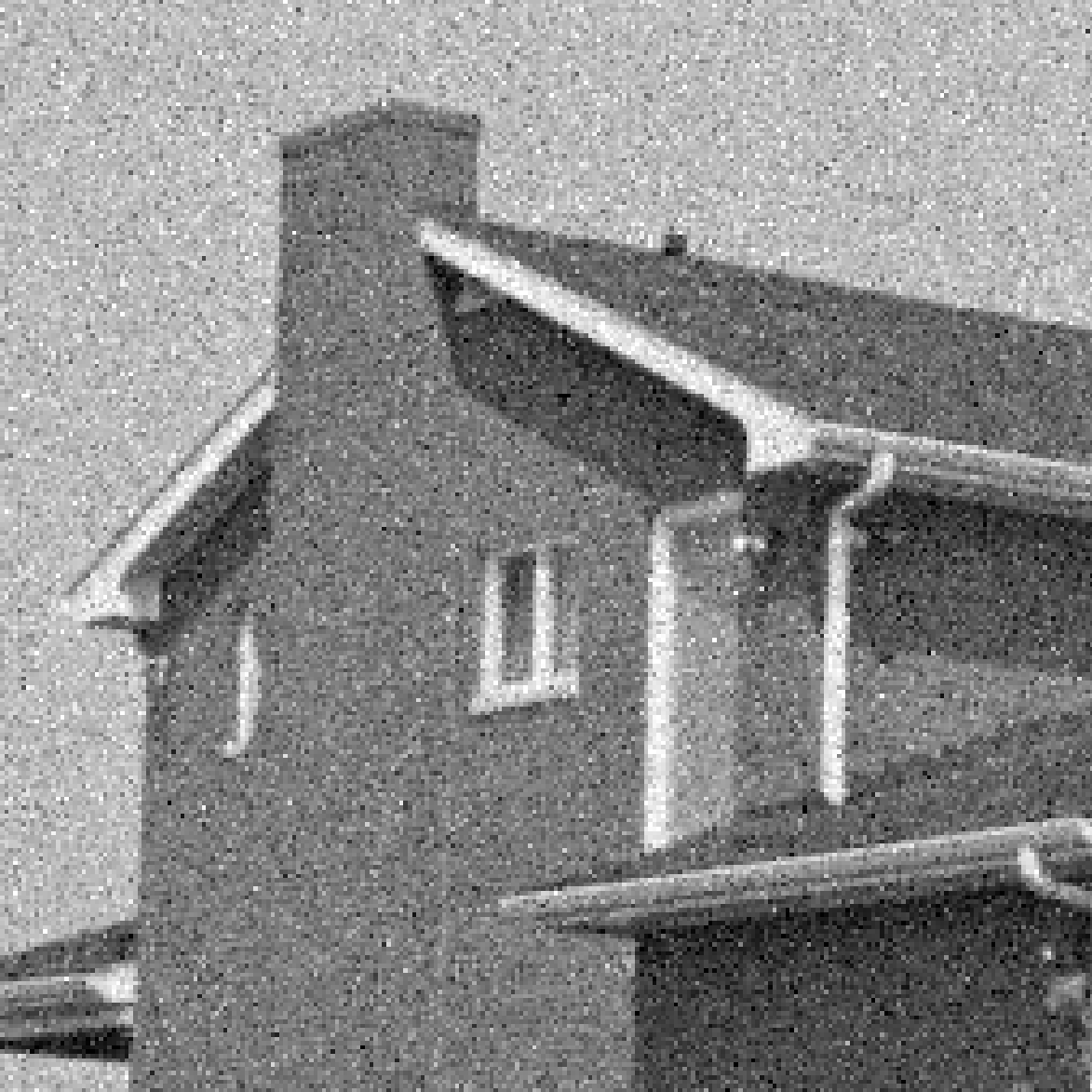}\\
		
		(a)   & (b)   \\
		\includegraphics[width=1.5in]{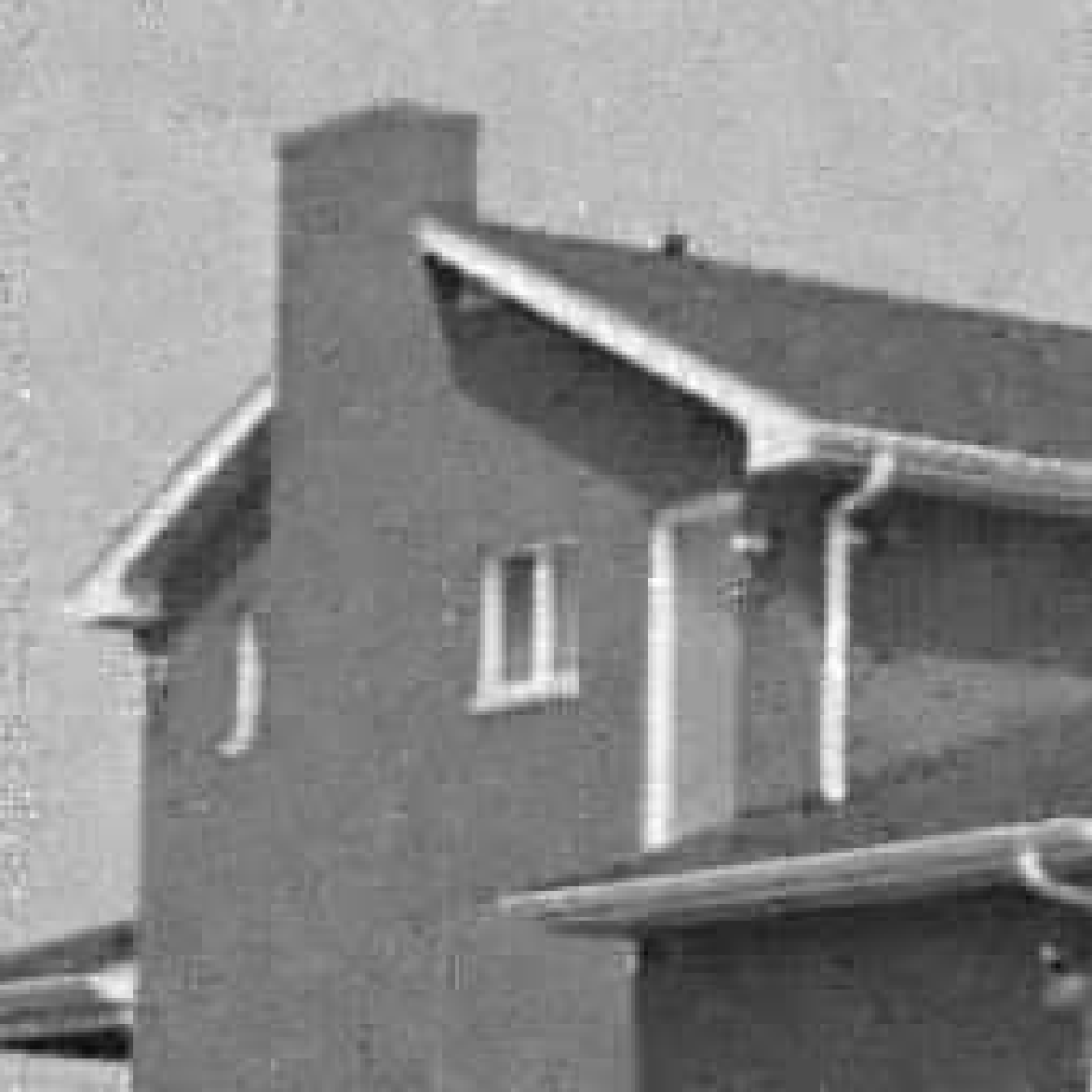}&
		\includegraphics[width=1.5in]{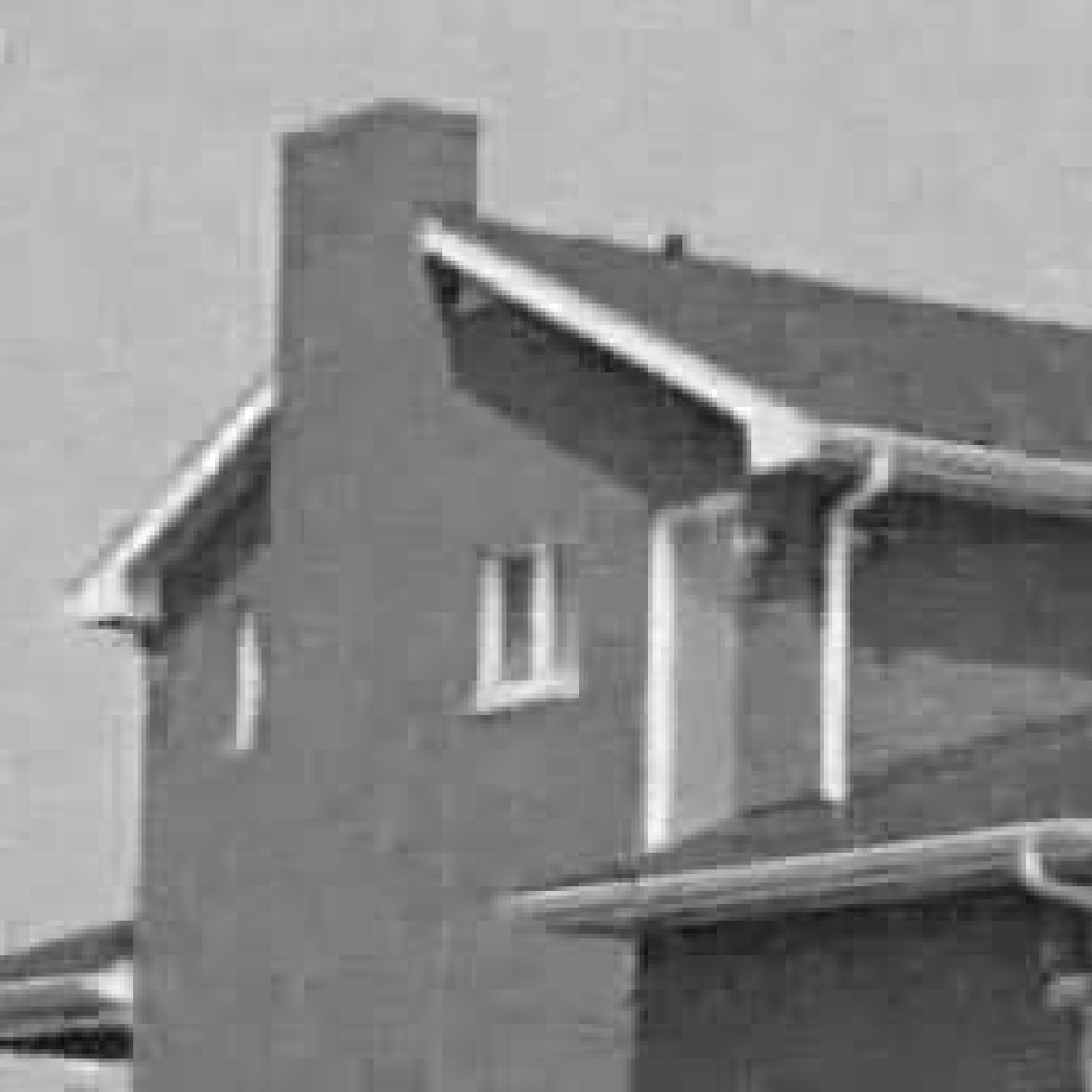}\\
		
		(c)  & (d)  \\
	\end{tabular}
	\caption{\small{Comparison of denoising performance of the K-SVD and $\ell_1$-K-SVD algorithms for Laplacian noise: (a) Ground-truth clean image; (b) Noisy input, $\text{PSNR}=22.16$ dB, $\text{SSIM}=0.36$; (c) Denoised image using K-SVD, $\text{PSNR}=29.78$ dB, $\text{SSIM}=0.73$; (d) Denoised image using $\ell_1$-K-SVD, $\text{PSNR}=30.34$ dB, $\text{SSIM}=0.81$. The values of $\lambda$ are $n_p$ are $1$ and $0.05$, respectively.}\vspace{-.2in}}
	\label{image_denoising_fig}
\end{figure}
\end{center}

We conduct image denoising experiments for additive Gaussian and Laplacian noise. The peak signal-to-noise ratio (PSNR) and structural similarity index (SSIM) of the output images obtained using the K-SVD and $\ell_1$-K-SVD algorithms are compared. In our experiments, dictionary learning is performed adaptively using patches taken from the noisy image \cite{elad3}. Patches of size $8\times 8$ are extracted from the noisy image, with an overlap of $4$ pixels in both directions. Thus, for an image of size $256\times 256$, $3969$ noisy patches are extracted to train the dictionaries that are chosen to be of size $64\times 128$. Both K-SVD and the proposed $\ell_1$-K-SVD algorithms are initialized with the noisy patches, and the iterations are repeated $10$ times \cite{elad3}. In the sparse coding stage of the K-SVD algorithm, we solve an optimization problem of the form  $\underset{\bold x_n}{\min}  \left\|  \bold x_n \right\|_0 \text{\,\,subject to\,\,}\left\|\bold y_n-\hat{D} \bold x_n \right\|_2 \leq 1.15\sigma$,
where $\sigma^2$ is the variance of the noise and $\hat{D}$ denotes the current estimate of the dictionary. The factor $1.15$ is fixed experimentally \cite{elad3}. The $\ell_1$-K-SVD update of the coefficients is obtained by solving (\ref{sparse_coding}) for each noisy patch $\bold y_n$, followed by hard-thresholding. The values of $\lambda_n$ are chosen to be equal for each patch, and denoted by $\lambda$. Hard-thresholding is performed by retaining a fraction $n_p$ of $K$ entries in each column of $X$. The values of $\lambda$ and $n_p$ depend on the type and strength of the noise. As a rule of thumb, one must choose a larger value of $\lambda$ and smaller value for $n_p$ as the value of $\sigma$ increases. The optimal values of these two parameters are obtained experimentally and reported in Table \ref{table_params}. The estimated clean patches are averaged to generate the overall denoised output. The PSNR and SSIM of the denoised images, for additive Laplacian and Gaussian noise of different levels $\sigma$, are reported in Table \ref{table_image_gauss_laplacian}. We observe that the $\ell_1$-K-SVD algorithm yields an output PSNR comparable with the K-SVD, and results in a higher value of SSIM over K-SVD, especially when $\sigma$ is large. For images corrupted by Gaussian noise with low value of $\sigma$, K-SVD results in a better denoising performance than $\ell_1$-K-SVD. However, as the noise strength $\sigma$ increases, $\ell_1$-K-SVD does a better job of preserving the image structure, as reflected in the higher SSIM values. When the image is corrupted by Laplacian noise, $\ell_1$-K-SVD yields slightly better PSNR over K-SVD, but the improvement in SSIM is in the range of $0.08-0.10$. For visual comparison, we show the denoised output images obtained using K-SVD and $\ell_1$-K-SVD in Fig. \ref{image_denoising_fig}, where the input image is contaminated by Laplacian noise, with PSNR of $22.16$ dB. The values of PSNR and SSIM of the denoised images obtained using the two algorithms are indicated in Fig. \ref{image_denoising_fig}. We observe that the gain in PSNR using $\ell_1$-K-SVD is slightly better than K-SVD, and the improvement in SSIM is close to $0.1$.

\vspace{-0.4in}
\section{Conclusions}
\label{sec_con}
We have developed a robust dictionary learning algorithm, referred to as the $\ell_1$-K-SVD, for minimizing the $\ell_1$ error on the data term and applied it for image denoising. The motivation behind using the $\ell_1$ metric as a measure of data fidelity is to achieve robustness to non-Gaussian noise and to alleviate the problem of over-smoothing texture and edges in images. The $\ell_1$-K-SVD algorithm updates the dictionary and the coefficients simultaneously, offering better flexibility and faster convergence compared to the RDL algorithm in \cite{Lu}. Experiments on synthesized data indicate the superiority of the proposed algorithm over its $\ell_2$-based counterpart, namely the K-SVD algorithm, in terms of ADR and the distance of the recovered dictionary to the ground-truth. We have also demonstrated the superiority of the $\ell_1$ metric for the case where the number of training examples is limited. Image denoising experiments showed that the $\ell_1$-K-SVD algorithm results in output images with slightly higher PSNR and SSIM values, at low input PSNRs, compared with the K-SVD algorithm. The $\ell_1$-K-SVD algorithm may also turn out to be useful in other applications such as image super-resolution.

\appendices
\ifCLASSOPTIONcaptionsoff
  \newpage
\fi

\end{document}